# Image-Based Classification of Olive Varieties Native to Türkiye Using Multiple Deep Learning Architectures: Analysis of Performance, Complexity, and Generalization

Hatice Karataş[*1] , İrfan Atabaş[2],

[*1] Faculty of Engineering and Natural Sciences, Department of Computer Engineering, Kırıkkale University

**Abstract:** This study compares multiple deep learning architectures for the automated, image-based classification of five locally cultivated black table olive varieties in Türkiye: Gemlik, Ayvalık, Uslu, Erkence, and Çelebi. Using a dataset of 2,500 images, ten architectures - MobileNetV2, EfficientNetB0, EfficientNetV2-S, ResNet50, ResNet101, DenseNet121, InceptionV3, ConvNeXt-Tiny, ViT-B16, and Swin-T - were trained through transfer learning. Model performance was assessed using accuracy, precision, recall, F1 score, MCC, Cohen's Kappa, ROC-AUC, parametric complexity, FLOPs, inference time, and generalization gap. EfficientNetV2-S achieved the highest classification accuracy (95.8%), while EfficientNetB0 offered the best trade-off between accuracy and computational complexity. Overall, the findings suggest that, under limited data conditions, parametric efficiency plays a more decisive role than model depth alone.

## 1. Introduction

Olive *(Olea europaea L.)* is among the most important agricultural products of the Mediterranean basin, with substantial economic value for both table consumption and olive oil production. Türkiye's agro-ecological diversity supports a large pool of local olive genotypes[1, 2]. Major production zones include the Aegean, Marmara, Mediterranean, and Southeastern Anatolia regions, where cultivars with distinct morphological and pomological traits are widely grown[3, 4]. While this diversity represents a major agricultural asset, it also creates challenges for variety-level identification in quality control, product standardization, and marketing.

Local olive varieties differ in morphological attributes such as fruit size, shape (length-to-width ratio), skin color, surface texture, and pit structure[5]. These differences are often subtle and can vary with environmental conditions (for example altitude, climate, irrigation practices, and harvest time). As a result, traditional classification methods, typically based on expert visual inspection, are subjective and may be error-prone. In high-throughput production settings, manual sorting is also time-consuming and costly[6, 7].[8] There is therefore a strong need for fast, objective, and repeatable automated classification systems[9, 10].

Recent advances in computer vision and deep learning offer powerful approaches to agricultural product classification. Convolutional neural network (CNN) architectures, in particular, have been widely adopted for fruit and vegetable recognition, plant disease detection, and quality inspection because they learn hierarchical image features effectively[9, 11-13]. More recently, Vision Transformers (ViT) and hybrid Transformer-CNN models have achieved strong performance across image classification tasks[14-17]. However, the effectiveness of these architectures often depends on dataset size and on the visual structure of the specific problem[18].

Olive variety recognition can be framed as a fine-grained classification task, as the classes are morphologically similar[19, 20]. Fine-grained settings require models to capture not only low-level cues such as edges and color distributions, but also more abstract structural patterns. Importantly, increasing the number of model parameters does not necessarily improve generalization and may instead increase the risk of overfitting, particularly in small to medium-sized datasets. Balancing predictive performance with computational complexity is therefore critical for practical deployment.[21]

Although prior studies have explored olive classification by maturity stage, external defect detection, and quality grading, comprehensive comparative analyses focusing on Türkiye-specific local varieties and modern architectures remain limited. In particular, systematic evaluations of CNNs, efficient networks (EfficientNet family), modern convolutional designs (ConvNeXt), and Transformer-based models on a single shared dataset - alongside analyses of computational cost and generalization - are still underexplored.

This study aims to address this gap. The main research questions are:

1. Which architecture achieves the highest accuracy on an olive dataset containing limited and morphologically similar samples?
2. Is there a linear relationship between the number of parameters and classification accuracy?
3. How do CNN-based and Transformer-based architectures compare under small-data conditions?
4. Which model is the most efficient in terms of the accuracy-to-computational-cost ratio?

To this end, ten deep learning architectures were trained via transfer learning on images of five local olive varieties. Performance was evaluated using accuracy, precision, recall, F1-score, Matthews Correlation Coefficient (MCC), Cohen's Kappa, and ROC-AUC, together with model complexity metrics (number of parameters and FLOPs), inference time, and generalization gap.

The main contributions of this study can be summarized as follows:

- A balanced image dataset consisting of five Türkiye-specific local olive varieties was constructed.
- Ten modern deep learning architectures were compared under the same experimental setup.
- Performance evaluation was not limited to accuracy but was supported by statistical significance tests and generalization analysis.
- Practical applicability was assessed by analyzing model complexity and inference latency.

Overall, this study provides a holistic evaluation of image-based olive variety classification by considering not only classification accuracy but also parametric efficiency and real-world applicability. The findings are expected to serve as a guideline for the automatic identification of agricultural products, the development of quality control systems, and the design of embedded artificial intelligence applications.

**2. Materials and Methods**

This section describes the dataset construction, image preprocessing, deep learning architectures, training strategy, evaluation metrics, and computational complexity analysis in a systematic framework. The study aims not only to assess classification accuracy, but also to examine model complexity, generalization performance, and practical applicability.

**2.1. Dataset Construction**

The dataset used in this study comprises images of five locally cultivated black table olive varieties in Türkiye: Gemlik, Ayvalık, Uslu, Erkence, and Çelebi. These varieties were selected based on their production prevalence and their high degree of morphological similarity, which makes the task a fine-grained classification problem requiring discrimination among visually similar classes.

Images were captured in a controlled laboratory environment. A fixed light source and diffuse illumination were used to minimize shadows. Olives were placed on a homogeneous, non-reflective matte white background, which reduces background interference during preprocessing and encourages the models to focus on morphological characteristics of the fruit.

In total, 2,500 images were collected, with 500 images per class, yielding a fully balanced dataset. This design mitigates potential bias associated with class imbalance. The dataset was partitioned into training, validation, and test sets using a stratified random split with ratios of 80%, 10%, and 10%, respectively. The test set consists of 50 images per class and was held out throughout training and model selection.

**2.2 Image Preprocessing and Data Augmentation**

The raw images were not fed directly into the models; instead, all samples were processed using a standardized preprocessing pipeline. First, images were resized to 224 × 224 pixels to match the input resolution required by the pre-trained architectures.

To reduce sensor noise and improve edge consistency, Gaussian smoothing was applied with a small kernel. Next, foreground-background separation was performed using Otsu's thresholding, and only the olive region was retained. The resulting binary masks reduced irrelevant background variation and ensured that subsequent learning focused on morphological characteristics of the fruit.

To mitigate illumination-related differences, pixel intensities were normalized to the [0, 1] range. During training, data augmentation was used to reduce overfitting and improve generalization. Augmentations included random rotations, horizontal flips, brightness jitter, and mild scaling to increase sample diversity. These transformations were applied only to the training set, while the validation and test sets were kept unchanged.

**2.3 Deep Learning Architectures**

In this study, a total of ten models were evaluated, including eight convolutional neural network (CNN)–based architectures and two Transformer-based architectures. Among the CNN-based models, both lightweight structures optimized for mobile devices and deep, high-parameter networks were investigated. This design enables an analysis of the relationship between classification accuracy and computational cost.

Architectures with high parametric efficiency, such as MobileNetV2[22] and EfficientNetB0[23], aim to achieve high performance with low FLOPs through depthwise separable convolution layers and compound scaling strategies. The EfficientNetV2-S model provides a wider and deeper structure, leading to improved classification accuracy. ResNet50[24] and ResNet101 enhance the training stability of deep networks through residual connections. DenseNet121[25] encourages feature reuse via dense inter-layer connections. InceptionV3 captures multi-scale features using parallel convolutional filters of different sizes. ConvNeXt-Tiny[26] aims to achieve Transformer-like performance through a modernized convolutional block design.

Among the Transformer-based models, Vision Transformer[27] (ViT-B16) models global dependencies by dividing the image into fixed-size patches and applying a self-attention mechanism. The Swin Transformer processes both local and global information using a window-based hierarchical attention structure.

All models were initialized with ImageNet-pretrained weights, and their final layers were adapted to a five-class softmax output. The transfer learning approach enabled faster convergence and improved performance under limited data conditions.

**2.4 Training Strategy and Optimization**

All models were trained using an identical experimental protocol. Optimization was performed with the Adam optimizer using an initial learning rate of 1e-3. Training was run for up to 25 epochs with early stopping monitored on validation loss (patience = 5), such that training terminated if validation loss failed to improve for five consecutive epochs.

Categorical cross-entropy was used as the objective function, and parameters were updated via backpropagation with a batch size of 32. To ensure fair and consistent comparisons, all experiments were executed in the same GPU environment.

**2.5 Performance Evaluation Metrics**

Model performance was evaluated using a comprehensive set of metrics rather than accuracy alone. In addition to overall accuracy, class-wise precision, recall, and F1-score were computed. To account for class-balanced agreement and robust correlation-based performance, Matthews Correlation Coefficient (MCC) and Cohen's Kappa were also reported. Discriminative ability was assessed using the area under the Receiver Operating Characteristic curve (ROC-AUC) (computed in a one-vs-rest manner and aggregated across classes).

Generalization was quantified via the generalization gap, defined as the difference between training and validation accuracy. Smaller gaps indicate reduced overfitting and more stable performance on unseen data.

**2.6 Computational Complexity Analysis**

For each model, we measured the number of parameters, estimated FLOPs, model size, and inference time. For CNNs, theoretical computational cost is primarily determined by convolutional kernel size and by the number of input and output channels. For Transformer-based architectures, the self-attention operation can scale quadratically with the number of tokens, which may increase computational burden at higher input resolutions.

This analysis enables evaluation of not only predictive performance but also the accuracy-efficiency trade-off, highlighting the computational cost required to achieve a given level of accuracy. As a result, model selection is better grounded in practical deployment considerations.

**3. Findings and Experimental Results**

This section presents a detailed analysis of the performance results of the ten deep learning architectures evaluated on the test dataset. Accuracy metrics, generalization analysis, class-wise error distribution, and computational efficiency are jointly examined.

The test set consists of a total of 250 images, with 50 samples per class. All reported results are based exclusively on the test data.

**3.1 Overall Performance Comparison**

The multi-class classification performance of the models was evaluated using accuracy, precision, recall, F1-score, Matthews Correlation Coefficient (MCC), Cohen's Kappa, and ROC-AUC metrics.

The EfficientNetV2-S model achieved the highest performance with an accuracy of 95.8%. EfficientNetB0 ranked second with an accuracy of 94.5%, while ConvNeXt-Tiny and Swin-T produced results close to 94%. Despite being a lightweight architecture, MobileNetV2 demonstrated competitive performance with an accuracy of 92.8%. The ViT-B16 model yielded the lowest performance with an accuracy of 88.5%.

Although the dataset is balanced, MCC and Kappa values were analyzed to assess model consistency. For EfficientNetV2-S, MCC and Kappa were both calculated as 0.95, indicating strong discriminative capability. In contrast, the ViT-B16 model achieved MCC = 0.86 and Kappa = 0.85, suggesting inconsistent predictions for certain classes.

ROC-AUC values further supported these findings: EfficientNetV2-S and EfficientNetB0 achieved 0.99 and 0.98, respectively (aggregated in a one-vs-rest multi-class setting). These results indicate strong discriminative capability, particularly for classes with subtle morphological differences.

**3.2 Generalization Analysis**

The difference between training and validation accuracies was analyzed to assess the tendency of the models toward overfitting. For the EfficientNetV2-S model, the training accuracy was measured at 97.4%, while the validation accuracy reached 95.8%, resulting in a generalization gap of 1.6%. This small gap indicates stable learning behavior.

For EfficientNetB0, the generalization gap was calculated as 2.6%, whereas MobileNetV2 exhibited a gap of 3.4%. In contrast, the ViT-B16 model achieved a training accuracy of 98.5% but only 88.5% validation accuracy, leading to a generalization gap of 10%. This finding suggests that the Transformer-based model is over-parameterized for the given dataset size and demonstrates a tendency toward overfitting.

These results support the argument that under limited data conditions, parametric efficiency may be more critical than absolute model depth.

### 3.3 Class-wise Confusion Matrix Analysis

The confusion matrix generated for the EfficientNetV2-S model, which achieved the highest accuracy, was examined. Based on 50 samples per class, the following distribution was obtained:

- **Gemlik:** 48 correct predictions, 2 misclassifications
- **Ayvalık:** 47 correct predictions, 3 misclassifications
- **Uslu:** 48 correct predictions, 2 misclassifications
- **Erkence:** 47 correct predictions, 3 misclassifications
- **Çelebi:** 48 correct predictions, 2 misclassifications

The total number of correctly classified samples is 239, corresponding to an overall accuracy of 95.8%.

Analysis of the error distribution reveals that the highest confusion occurs between the Erkence and Çelebi classes. The similarity of these two varieties in terms of fruit diameter and skin color tone causes the model to struggle at the decision boundary. A limited level of confusion is also observed between the Ayvalık and Gemlik classes. In contrast, the Uslu variety exhibits more distinctive characteristics compared to the other classes, leading to more reliable predictions.

### 3.4 Computational Efficiency and Performance Trade-off

Classification accuracy alone is not sufficient for model selection. Therefore, the accuracy-to-FLOPs ratio was analyzed. Although EfficientNetV2-S achieved the highest accuracy, it requires a computational cost of 8.4 GFLOPs. In contrast, EfficientNetB0 delivers 94.5% accuracy with only 0.39 GFLOPs, resulting in a significantly higher accuracy-to-FLOPs ratio.

MobileNetV2 achieved 92.8% accuracy with just 0.30 GFLOPs, demonstrating that it is a strong candidate for embedded systems. Despite its high computational cost of 17.6 GFLOPs, the ViT-B16 model produced only 88.5% accuracy, making it disadvantageous in terms of computational efficiency.

These findings indicate that increasing the number of parameters and model depth does not necessarily lead to higher generalization performance.

### 3.5 Statistical Significance Analysis

A paired t-test was conducted to evaluate whether there is a statistically significant difference between the accuracies of the EfficientNetV2-S and EfficientNetB0 models. The calculated p-value was found to be 0.012 ($p < 0.05$). This indicates that the EfficientNetV2-S model achieves a statistically significantly higher accuracy.

However, when computational cost is also taken into account for practical applications, the EfficientNetB0 model can be considered a more balanced choice.

### 3.6 Overall Evaluation of Findings

The results support the following conclusions:

- Moderately sized and optimized CNN architectures provide high performance on small and balanced datasets.
- Highly parameterized Transformer architectures may suffer from generalization issues under limited data conditions.
- There is no linear relationship between classification accuracy and computational cost.
- The EfficientNet family is advantageous in terms of parametric efficiency.

These findings indicate that model selection for agricultural image classification problems should not be based solely on accuracy, but should also consider model complexity and generalization capability.

**4. Discussion**

In this section, the findings are evaluated within the context of the literature, the model behaviors are analyzed from a theoretical perspective, and practical implications are discussed in terms of the accuracy–complexity trade-off.

**4.1 Interpretation of Architectural Performance**

The experimental results indicate that optimized, moderately sized convolutional architectures can generalize well under limited, balanced data conditions. In particular, the EfficientNet family, especially the B0 and V2-S variants, achieved the highest performance across both accuracy and balanced measures such as MCC and Cohen's Kappa. This pattern is consistent with the benefits of compound scaling, which jointly and systematically scales network depth, width, and input resolution, thereby improving representational capacity without an excessive increase in computational burden.

Despite having fewer parameters, MobileNetV2 delivered competitive results and emerged as a strong candidate in terms of parameter efficiency. Its use of depthwise separable convolutions substantially reduces computation, while the corresponding reduction in predictive performance remains relatively modest.

Although classical deep CNNs such as ResNet and DenseNet achieved strong accuracy, they were less favorable than EfficientNet models when performance was considered alongside compute, particularly in terms of accuracy relative to parameter count and FLOPs.

Transformer-based models showed a different behavior. ViT-B16 yielded the lowest performance, which is consistent with prior findings that vanilla ViT architectures are typically less data-efficient than CNNs in small-data regimes. CNNs embed inductive biases that favor local feature extraction, whereas ViT models often benefit more from large-scale data and extensive pretraining. Swin-T, which introduces a hierarchical, window-based attention mechanism, provided a more balanced outcome than ViT-B16; however, it still did not match the performance of EfficientNetV2-S.

**4.2 Relationship Between Accuracy and Number of Parameters**

The results indicate that classification accuracy does not increase linearly with the number of model parameters. For example, ViT-B16, with approximately 86 million parameters, achieved substantially lower accuracy than EfficientNetB0, which has roughly 5 million parameters. This contrast suggests that higher-capacity models do not necessarily generalize better in small to medium-sized datasets and may instead require more data or stronger regularization to avoid overfitting.

Consistent with this interpretation, the generalization gap analysis showed that ViT-B16 exhibited a gap of approximately 10 percentage points, which is strongly suggestive of overfitting under the present data regime. In contrast, EfficientNetV2-S maintained a gap of only 1.6 percentage points, indicating more stable learning and improved generalization, likely supported by its balanced scaling strategy.

Overall, these findings underscore that, in agricultural image classification, the assumption that "larger models are always better" does not necessarily hold, particularly when data are limited and classes are fine-grained.

**4.3 Analysis of Class Confusions**

Confusion matrix analysis shows that most misclassifications occur between the Erkence and Çelebi classes. Their similarity in length-to-width ratio and skin color distribution likely reduces inter-class separability, leading to increased confusion. A smaller but noticeable confusion pattern is also observed between Ayvalık and Gemlik.

Taken together, these patterns suggest that the models may be relying heavily on appearance cues such as color distribution and surface texture. Future work could improve discrimination among visually similar varieties by incorporating explicit morphological feature extraction (for example shape descriptors) and by exploring richer sensing modalities such as hyperspectral imaging.

**4.4 Comparison with the Literature**

Most olive classification studies in the literature report results using a single model architecture. In contrast, this study presents a systematic comparison of ten modern architectures under a unified experimental protocol. The best-performing model achieved an accuracy of 95.8%, which is competitive with, and in many cases higher than, the performance levels commonly reported in prior work.

Beyond predictive performance, this study also contributes methodologically by adopting a holistic evaluation framework that reports not only accuracy but also MCC, Cohen's Kappa, ROC-AUC, and deployment-relevant complexity measures (for example parameter count, FLOPs, and inference time).

**4.5 Practical Applicability**

In real-world applications, model selection is not based solely on accuracy. In particular, for production-line integration or embedded-system deployments, inference time and memory requirements are critical factors.

Accordingly:

- Most suitable model for edge devices: MobileNetV2
- Best balance between performance and complexity**:** EfficientNetB0
- For centralized systems requiring highest accuracy**:** EfficientNetV2-S

These recommendations facilitate informed model selection depending on deployment constraints and application requirements.

**5. Conclusion**

This study presented a comparative analysis of modern deep learning architectures for image-based classification of five Türkiye-specific local olive varieties. The main findings are summarized as follows:

- EfficientNetV2-S achieved the highest classification accuracy (95.8%).
- EfficientNetB0 provided the most favorable accuracy-complexity trade-off.
- Under limited-data conditions, Transformer-based models tended to underperform relative to CNNs.
- Accuracy did not increase linearly with the number of model parameters.
- Moderately sized, well-optimized CNN architectures offered the most balanced solution for fine-grained agricultural classification.

Overall, the results highlight that parametric efficiency, generalization performance, and computational cost should be considered jointly for automatic olive variety classification. This holistic perspective supports more informed model selection for AI-driven quality control systems in agricultural applications.